\documentclass[10pt,twocolumn,letterpaper]{article}

 \usepackage{iccv}              
\usepackage{relsize}
\usepackage{multirow}
\usepackage[accsupp]{axessibility}
\definecolor{iccvblue}{rgb}{0.21,0.49,0.74}
\usepackage[pagebackref,breaklinks,colorlinks,allcolors=iccvblue]{hyperref}

\def\paperID{7076} 
\def\confName{ICCV}
\def\confYear{2025}

\title{M-SpecGene: Generalized Foundation Model for RGBT Multispectral Vision}

\author{Kailai Zhou\textsuperscript{1,2}, Fuqiang Yang\textsuperscript{1}, Shixian Wang\textsuperscript{1}, Bihan Wen\textsuperscript{2}, Chongde Zi\textsuperscript{1},\\ Linsen Chen\textsuperscript{1}\footnotemark[1], Qiu Shen\textsuperscript{1}, Xun Cao\textsuperscript{1}\footnotemark[1] \\
\textsuperscript{1}Nanjing University, Nanjing, China \ \  \textsuperscript{2}Nanyang Technological University, Singapore\\
{\tt\small calayzhou@smail.nju.edu.cn} \ \ {\tt\small \{chenls, caoxun\}@nju.edu.cn}}

\begin{document}
\maketitle

\begin{abstract}
RGB-Thermal (RGBT) multispectral vision is essential for robust perception in complex environments. Most RGBT tasks follow a case-by-case research paradigm, relying on manually customized models to learn task-oriented representations. Nevertheless, this paradigm is inherently constrained by artificial inductive bias, modality bias, and data bottleneck. To address these limitations, we make the initial attempt to build a Generalized RGBT MultiSpectral foundation model (M-SpecGene), which aims to learn modality-invariant representations from large-scale broad data in a self-supervised manner. M-SpecGene  provides new insights into multispectral fusion  and integrates prior case-by-case studies into a unified paradigm. Considering the unique characteristic of information imbalance in RGBT data, we introduce the Cross-Modality Structural Sparsity (CMSS) metric to quantify the information density across two modalities. Then we develop the GMM-CMSS progressive masking strategy to facilitate a flexible, easy-to-hard, and object-centric pre-training process. Comprehensive experiments validate M-SpecGene's generalizability across eleven datasets for four RGBT downstream tasks. The code will be available at \url{https://github.com/CalayZhou/M-SpecGene}.





\end{abstract}    
\section{Introduction}
\label{sec:intro}
\begin{figure}[t]
	\centering
	\includegraphics[width=1.0\linewidth]{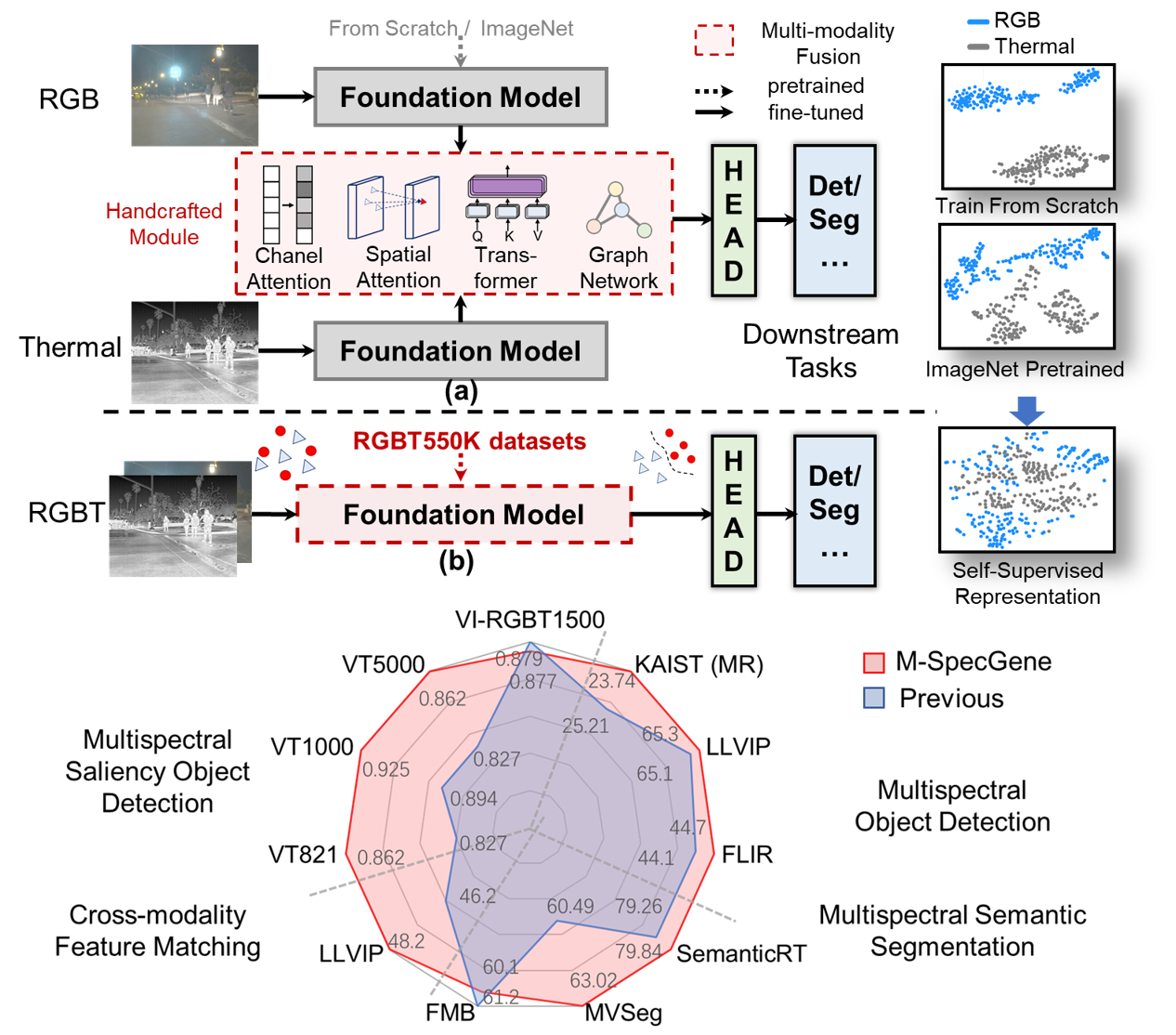}
	\vspace{-2.4em}
	\caption{(a) Manually customized models: task-oriented representations are learned under a case-by-case research paradigm.  (b) Generalized RGBT multispectral foundation model aims to learn modality-invariant representations by self-supervised learning. The t-SNE visualization of RGB and thermal features indicates M-SpecGene  achieves superior cross-modality alignment.  }
	\label{fig:introduction}
 \vspace{-1.4em}
\end{figure}


RGB sensors alone struggle to handle complex environmental conditions, including smog, low light, and high dynamic range scenarios. RGBT multispectral vision, with its all-weather, round-the-clock sensing capabilities, has emerged as a crucial technology in fields like autonomous driving, military defense, remote sensing, and industrial inspection.

Currently, most RGBT downstream tasks follow a case-by-case research paradigm. For a given task, task-oriented representations are learned via fully supervised learning on small, task-specific datasets, often using models pretrained on ImageNet or trained from scratch. As illustrated in Fig.~\ref{fig:introduction}(a), existing methods commonly use two-stream branches to extract features from both RGB and thermal images, incorporating complex handcrafted modules in the intermediate feature space, such as channel attention \cite{zhou2020improving}, spatial attention \cite{yun2022infusion}, Transformer \cite{qingyun2021cross}, and graph network \cite{song2022multiple}. However, this case-by-case paradigm has several limitations:
\textbf{1) Artificial inductive bias:} Task-oriented, manually customized models, being optimized for a given task, are effective for that task but may lead to suboptimal results on others, thereby restricting both the scalability of the designed model and the generalizability  of the learned representations. \textbf{2) Modality bias:} Due to inherent differences between RGB and thermal modalities, initializing the thermal branch with the ImageNet pretrained model inevitably introduces modality bias. This bias can potentially impair the encoded prior knowledge and result in suboptimal feature representations for the thermal modality. \textbf{3) Data bottleneck:} RGBT multispectral images are harder to obtain than single RGB images, and high-quality manual annotation for large datasets is costly and time-intensive.




Recently, foundation models, with their capacity to encode extensive knowledge \cite{awais2023foundational}, offer a potential solution to above limitations. As shown in Fig.~\ref{fig:introduction}(b), we make an initial attempt to transform manually customized models into a generalized multispectral foundation model named M-SpecGene, which aims to explore a new RGBT fusion paradigm that learns modality-invariant representations in a self-supervised manner, therefore eliminating the need for handcrafted modules and facilitating multi-modality feature fusion in a simple yet effective way. However, the self-supervised pre-training of generalized multispectral foundation model is challenging, due to the lack of large-scale datasets and the inherent information imbalance in RGBT data. In contrast to RGB images, thermal images lack rich textures, colors, and fine details. Moreover, significant differences in imaging mechanisms introduce asymmetry in information density between the two modalities. Additionally, RGBT datasets are not object-centric like ImageNet \cite{deng2009imagenet}; instead, they tend to include smaller, less salient objects with dispersed and uneven information distribution. 

To address above problems, M-SpecGene employs a Siamese architecture and a progressive masking strategy  to promote consistent representations in latent space. Leveraging the unique correlations within multispectral images, we introduce cross-modality structural sparsity to quantify information density between two modalities. Then we develop a Gaussian Mixture Model (GMM) to fit the overall CMSS distribution of the whole pre-training datasets, enabling a flexible, modality-balanced masking strategy that progresses from easier to more difficult learning stages. Our GMM-CMSS progressive masking strategy alleviates the impact of information imbalance in self-supervised pre-training, enhancing the encoder's ability to focus on consistent, modality-invariant, and object-centric representations.






M-SpecGene provides new insights into the RGBT fusion paradigm and offers the following advantages: \textbf{1) Simplified model design:} A single foundation model can effectively represent both RGB and thermal modalities, eliminating the need for complex handcrafted modules and facilitating the adaptation of single-modality RGB methods to RGBT two-modality tasks. \textbf{2) Generalized representation:} Self-supervised pre-training on large-scale data enables M-SpecGene to learn a versatile representation that overcomes limitations associated with artificial inductive and modality biases, making it adaptable to a diverse range of downstream tasks. \textbf{3) Enhanced data utilization:} M-SpecGene fully integrates self-supervised pre-training data from existing RGBT tasks without the need for human annotations. Our contributions are as follows:

$\bullet$ We make the first attempt to build a multispectral foundation model, M-SpecGene, exploring a new RGBT fusion paradigm that eliminates the need for handcrafted modules. 

$\bullet$ A high-quality, large-scale dataset, RGBT550K is carefully constructed for self-supervised pre-training.





$\bullet$ Considering the unique characteristic of RGBT datasets, we introduce a GMM-CMSS progressive masking strategy to mitigate the impact of information imbalance.



$\bullet$ M-SpecGene integrates prior case-by-case studies into a unified paradigm and demonstrates strong generalizability across eleven datasets for four RGBT downstream tasks.  
 
 



\section{Related Work}

\subsection{Task-Oriented RGBT Multispectral Vision}


We first make an overview of the related RGBT multispectral vision tasks.
\textbf{a) Multispectral Object Detection:}  Previous methods can be divided into three categories: 1) Early fusion at the image level. 2) Halfway fusion at the feature level. 3) Late fusion in a post-process manner. Halfway fusion has emerged as a primary focus, involving  an interaction module across modalities, such as channel attention \cite{zhou2020improving}, spatial attention \cite{yun2022infusion,an2022effectiveness,zhang2021guided}, and Transformer \cite{qingyun2021cross,lee2024crossformer,shen2024icafusion}. 
\textbf{b) Multispectral Semantic Segmentation:} Early studies adopt straightforward strategies, such as concatenating RGB and thermal features \cite{ha2017mfnet} or integrating thermal features into the RGB encoder \cite{deng2021feanet, zhou2021mffenet}. Recent investigations explore weighted attention-based fusion strategies to achieve robust cross-modality  fusion, utilizing techniques such as multi-scale spatial and channel context modules \cite{zhang2021abmdrnet}, explicit complement modeling framework \cite{ji2023semanticrt}, edge-aware guidance fusion \cite{zhou2022edge}, and spatio-temporal context integration \cite{ji2023multispectral}. 
\textbf{c) RGBT Cross-modality Feature Matching:} 
Modality-invariant representation plays a crucial role in cross-modality feature matching. Traditional handcrafted methods \cite{li2019rift} design reliable filters that exhibit certain robustness to modality differences, while recent deep learning methods \cite{deng2022redfeat} leverage loss functions to supervise the extraction of features. Nevertheless, existing methods suffer from limited generalization and robustness. \textbf{d) Multispectral Salient Object Detection:} 
Compared to semantic segmentation, saliency object detection faces challenges such as background complexity and contextual understanding. Thus, technologies such as the manifold ranking algorithm \cite{wang2018rgb}, multi-interaction block \cite{tu2020multi}, and multiple graph affinity interactive network \cite{song2022multiple} are proposed. 

In conclusion, previous RGBT downstream tasks primarily follow a case-by-case research paradigm. In this paper, we explore the transformation of multispectral fusion paradigm from the perspective of foundation model.


\subsection{Spectral Foundation Model}
Foundation models are initially pretrained on large-scale broad data in a self-supervised manner, and can be adapted (e.g., fine-tuned) for a wide range of downstream tasks \cite{awais2023foundational}. Foundation models driven by self-supervised learning for specialized data types have emerged in various areas,  such as SARATR-X \cite{LiSARATRX25} for synthetic aperture radar, InfMAE  \cite{liu2025infmae} for infrared images, and EVA-X \cite{yao2024eva} for X-ray images. Research on spectral foundation models mainly focuses on hyperspectral images in remote sensing, including SpectralGPT \cite{hong2024spectralgpt} and HyperSIGMA \cite{wang2024hypersigma}. Currently, there is a lack of research into the RGBT multispectral foundation model. A recent relevant work, UniRGB-IR \cite{yuan2024unirgb}, 
utilizes ViT-B as the pretrained foundation model and dynamically introduces richer RGB-IR features into the RGB-based pretrained model. Nevertheless, UniRGB-IR still requires the handcrafted fusion module and the adapter tuning design may not make adequate integration of two modalities. We make an initial attempt to develop multispectral foundation model, aiming to eliminate handcrafted modules by fully exploit large-scale RGBT data in a self-supervised manner.







\subsection{Information-aware Masking Strategy}
Compared to ImageNet \cite{deng2009imagenet}, RGBT datasets exhibit a distinct characteristic of information imbalance.  One solution involves an information-aware masking strategy, which aims to optimally choose what parts of the image to mask based on the informational value. For thermal images, InfMAE \cite{liu2025infmae} implements information-aware masking based on gray values. For RGB images, previous methods rely on teacher-student framework \cite{wang2023hard, kakogeorgiou2022hide}, semantic information learned by ViT \cite{li2022semmae}, CLIP \cite{hou2022milan} or segmentation task pre-training \cite{wu2022object} to measure information density distribution.  However, these methods often necessitate extra components or incur higher computational costs.  Furthermore, it should be noted that single-modality-based methods are difficult to adapt to multispectral images directly. We contend that the unique correlations between the two modalities can be leveraged to offer valuable clues for advanced information-aware masking.


\section{RGBT550K Dataset}







To pretrain a multispectral foundation model with robust generalization capabilities, we exert our utmost efforts to make a comprehensive collection of available RGBT datasets, resulting in three million RGBT samples (termed RGBT3M) drawn from 41 datasets and 10 multispectral tasks. Although RGBT3M offers substantial image quantity, we argue that diversity and quality are more critical. The RGBT3M dataset has several limitations: 1) Imbalance across datasets: RGBT detection and segmentation datasets \cite{zhang2020multispectral,liu2023multi}, typically contain fewer than 10,000 samples, while RGBT tracking datasets \cite{li2021lasher} often exceed 100,000 samples. 2) Temporal redundancy: Although tracking datasets contain hundreds of thousands of samples, they cover only a few hundred unique scenarios, leading to significant temporal redundancy; 3) Low image quality: Many datasets are captured in challenging conditions, such as nighttime or rainy scenes, resulting in lower imaging quality.

\begin{figure}[t]
	\centering
	\includegraphics[width=1.0\linewidth]{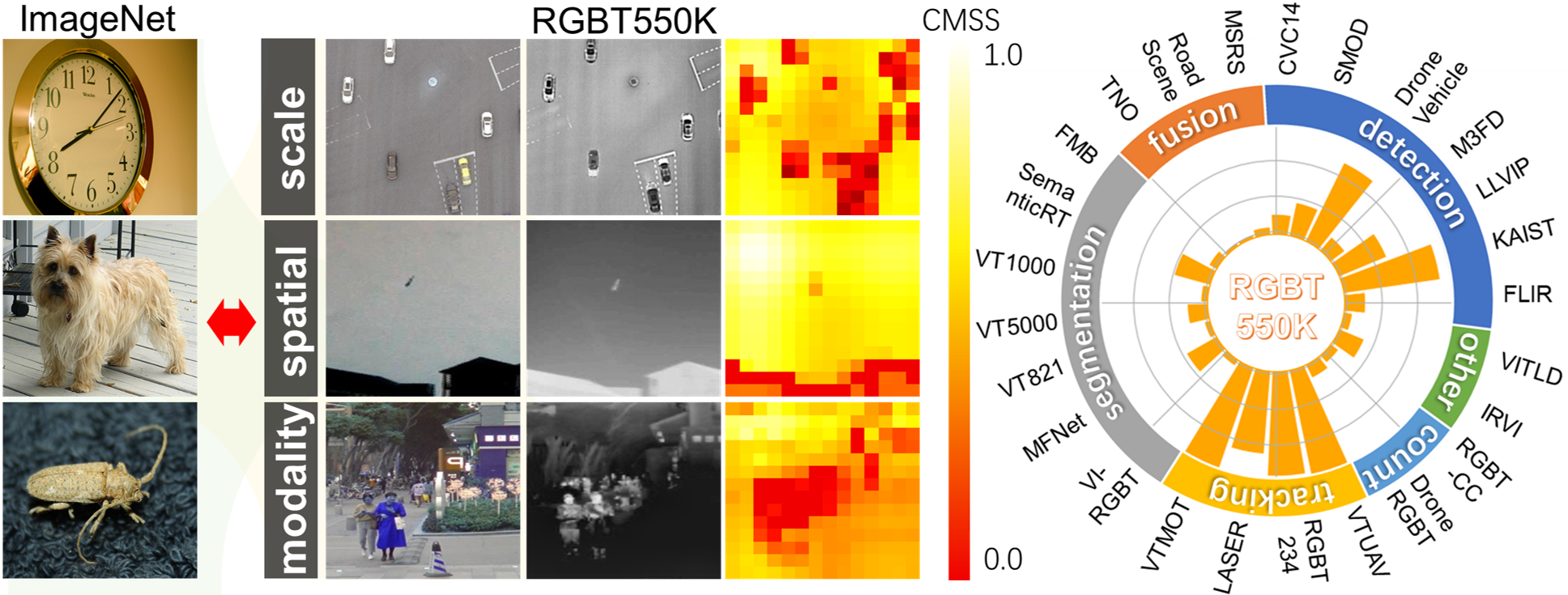}
	\vspace{-2.0em}
    \caption{ RGBT550K consists of diverse resources, it exhibits an imbalanced information distribution compared to ImageNet. }
	\label{fig:dataset}
 \vspace{-1.2em}
\end{figure}

Thus, we refine the RGBT3M through the following steps: 1) Ensuring dataset balance: We prevent any single dataset to dominate an excessive proportion. 2) Removing redundancy: Temporal sampling is applied to RGBT video datasets to eliminate highly similar frames. 3) Evaluating image quality: Using objective metrics, we find that SSIM \cite{wang2004image} is an effective measure of RGBT image quality. We remove samples with SSIM values below 0.80, as these images generally lack sufficient object information or are of poor quality. As shown in Fig.~\ref{fig:dataset}, our meticulous preprocessing yields RGBT550K, a comprehensive dataset comprising 548,238 high-quality samples. It encompasses diverse scenarios, tasks, lighting conditions, resolutions, and object categories, providing a solid foundation for the self-supervised pre-training of the multispectral foundation model. Further details can be found in the appendix.

\section{Method}


As shown in Fig.~\ref{fig:method}(a), our M-SpecGene adopts a Siamese-based architecture based on masked autoencoders \cite{he2022masked} for cross-modality self-supervised learning.  It begins with the GMM-CMSS progressive masking strategy, which dynamically selects masked patches based on information density. The complementary masked RGB and thermal patches are processed with a shared-weight ViT \cite{dosovitskiy2020image} encoder, a cross-attention layer is then employed to facilitate the propagation of complementary information in latent space. Finally, two modality-specific decoders with self-attention layers reconstruct the masked pixels for the RGB and thermal modalities independently. The Siamese-based architecture encourages both modalities to produce consistent representations. After self-supervised pre-training, we adopt the M-SpecGene ViT encoder for fine-tuning on downstream tasks, which will be explained in detail in Sec.~\ref{sec:fine-tune}.

The GMM-CMSS progressive masking strategy consists of three steps: 1) Given the uneven information distribution in RGBT datasets, we compute the CMSS metric for each RGBT image pair to quantify information density. 2) We employ Gaussian mixture modeling to estimate the overall CMSS distribution, which serves as a guide for subsequent information-aware masking. 3) A sampling function is designed based on GMM to implement the progressive masking strategy. With these steps, unmasked patches gradually move from foreground to background during pre-training.







\subsection{Cross-modality Structural Sparsity}
Fig.~\ref{fig:dataset} shows a prominent characteristic of RGBT datasets is their pronounced information imbalance, reflected in the uneven distribution of object scales, spatial and modality information density. Unlike ImageNet \cite{deng2009imagenet}, where objects are typically centered and occupy a larger portion of the image, RGBT datasets are not object-centered; they tend to contain smaller, less prominent objects with uneven spatial distribution. Additionally, differences in imaging mechanisms lead to modality imbalance \cite{zhou2020improving}, which means asymmetric information density between RGB and thermal modalities under varying conditions. Consequently, the random masking strategy used in MAE \cite{he2022masked} may disproportionately focus on information-sparse regions, undermining effective self-supervised learning. Therefore, we aim to develop an adaptive masking strategy based on the measurement of information density across modalities. Specifically, we divide RGB and thermal images into $p \times p$ non-overlapping patch embeddings, denoted as $A_{rgb}=\left\{a_i\right\}_{i=1}^{p \times p}$, $B_{t}=\left\{b_i\right\}_{i=1}^{p \times p}$, respectively. Here, $a_i, b_i \in \mathbb{R}^{768}$ are feature vectors of the $i$-th patch embeddings. For each patch embedding pair ($a$, $b$), we define cross-modality structural sparsity as follows:

\begin{figure}[t]
	\centering
	\includegraphics[width=0.95\linewidth]{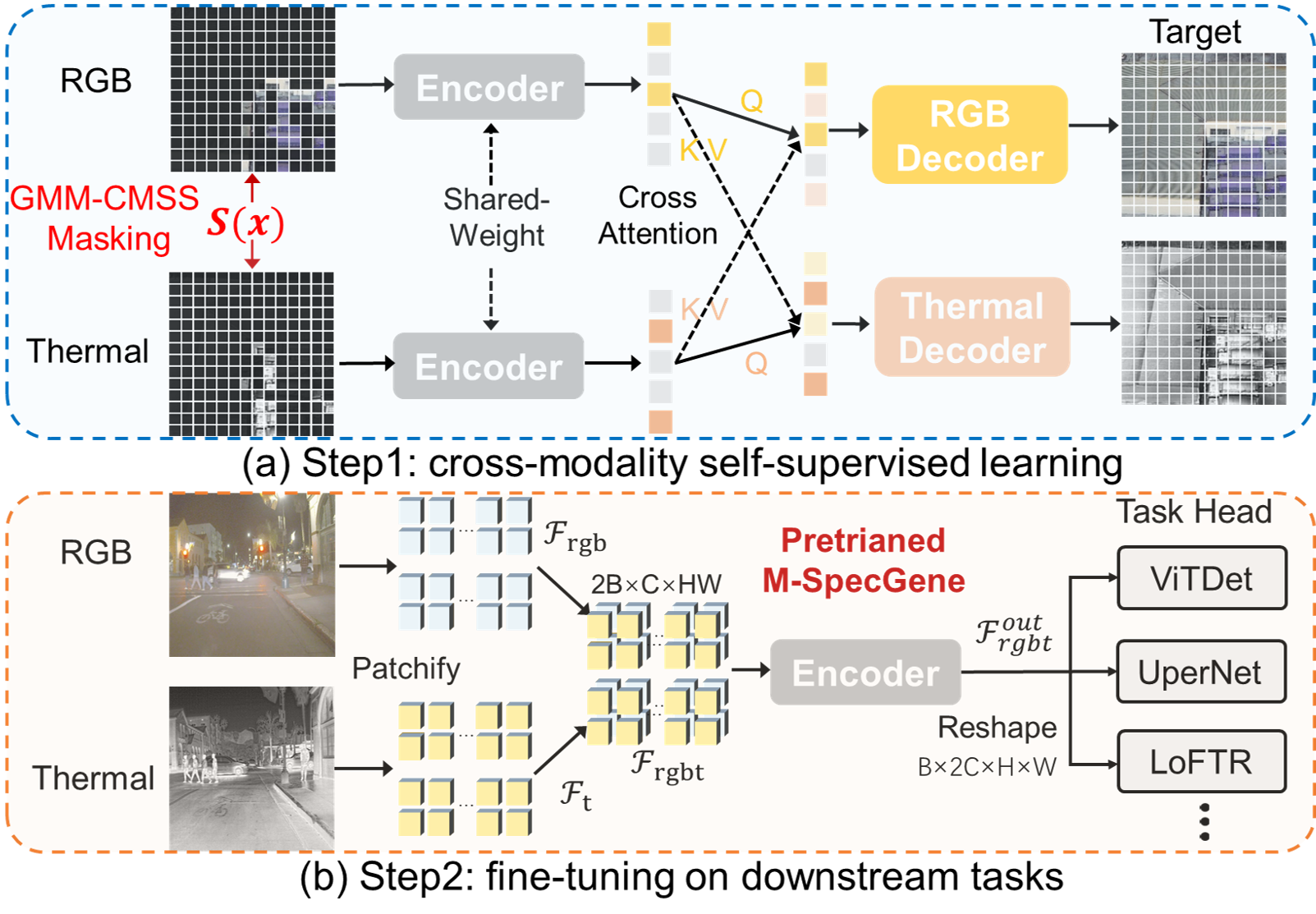}
	\vspace{-1.0em}
	\caption{(a) The self-supervised pre-training of M-SpecGene. (b) The fine-tuning of M-SpecGene on downstream tasks. }
    \label{fig:method}
 \vspace{-1.2em}
\end{figure}



\vspace{-0.5em}
\begin{equation}
\mathnormal{m}=\mathnormal{CMSS}(a, b)=\frac{1+<\frac{a}{|a|}, \frac{b}{|b|}>}{2 \sigma_a^2 \sigma_b^2}
\end{equation}
\noindent where the numerator represents the cosine similarity between RGB and thermal patch embeddings. The denominator consists of the structural variances of $a$ and $b$. To facilitate post-processing, the value of $m$ is normalized to the range $[0, 1]$. Fig.~\ref{fig:dataset} shows that in low information density regions (e.g., sky), patch embedding pairs ($a$, $b$) exhibit high similarity and low structural variance, resulting in relatively high CMSS value. Conversely, in high-information density regions (e.g., pedestrians), ($a$, $b$) exhibit greater differences, yielding lower similarity but higher structural variance. Consequently, CMSS tends to have lower values in regions with rich semantic context.  Thus, we employ the CMSS as a simple but effective metric to evaluate the information density across RGBT patch embedding pairs.


\subsection{CMSS Gaussian Mixture Modeling}
For the whole pre-training dataset comprising  $N$ image pairs, where each image pair contains  $p \times p$ patch embeddings, the overall CMSS distribution can be denoted as $ \textbf{m} =\left\{m_i\right\}_{i=1}^{N \times p \times p}$. The primary problem is to develop an effective masking strategy based on this overall CMSS distribution $\textbf{m}$. To address this, we first apply a Gaussian mixture model to estimate the whole CMSS distribution $\textbf{m}$ via maximum likelihood. After estimating the $ \textbf{m}$ with Gaussian mixture model, we dynamically adjust masked patches based on the Gaussian model associated with specific CMSS distribution intervals. We model the observed CMSS  $m$ for the patch embedding $(a, b)$ from the underlying distribution \textbf{m} as:

\vspace{-0.3em}
\begin{equation}
\setlength{\abovedisplayshortskip}{-5pt}
p(m)=\sum_{k=1}^K \pi_k \mathcal{N}\left(m \mid \mu_k, \Sigma_k\right)
\end{equation}
here, $p(m)$ represents the CMSS probability density function to be estimated by Gaussian mixture model; $K$ denotes the number of Gaussian components, which is set to 3 by default; $\pi_k$ is the weight of the $k$-th Gaussian component $\mathcal{N}\left(m \mid \mu_k, \Sigma_k\right)$ with mean $\mu_k$ and variance $\Sigma_k$. Calculating CMSS metrics for the entire pre-training dataset at once is computationally expensive. Moreover, the trainable linear projection parameters are continually updated during pre-training. Consequently, we aim to dynamically update the  Gaussian mixture model estimation, synchronized with the pre-training process on an epoch-by-epoch basis. During each pre-training iteration, we calculate $B \times p \times p$ CMSS samples, denoted as $\textbf{m}_{iter} =\left\{ m_i\right\}_{i=1}^{B\times p\times p}$ for $B$ image pairs in each  batch. In the estimation step, the posterior probability of each CMSS sample $m_i$ belonging to the $k$-th Gaussian model is estimated as follows:

\begin{equation}
\alpha_{i k}=\frac{\pi_k \mathcal{N}\left(m_i \mid \mu_k, \Sigma_k\right)}{\sum_{i=1}^{K} \pi_k \mathcal{N}\left(m_i \mid \mu_k, \Sigma_k\right)}
\end{equation}

Using the posterior probability $\alpha_{i k}$, we then update the parameters of the Gaussian mixture model $\left\{  \mu_{k}, \Sigma_k, \pi_k \right\}$ in the maximization step:


 \vspace{-1.2em}
\begin{equation}
\mathlarger{
\left\{\begin{array}{c}\mu_k=\frac{\sum_{i=1}^{B \times p \times p} \alpha_{i k} m_i}{\sum_{i=1}^{B \times p \times p} \alpha_{i k}} \\[10pt] \Sigma_k=\frac{\sum_{i=1}^{B \times p \times p} \alpha_{i k}\left(m_i-\mu_k\right)\left(m_i-\mu_k\right)^T}{\sum_{i=1}^{B \times p \times p} \alpha_{i k}} \\[10pt] \pi_k=\frac{\sum_{i=1}^{B \times p \times p} \alpha_{i k}}{B \times p \times p}\end{array}\right.}
\end{equation}

Following these steps, we iteratively update the Gaussian mixture model parameters $\left\{ \mu_{k}, \Sigma_k, \pi_k \right\}$ at each pre-training iteration to approximate the CMSS probability density function $p(m)$. Our observations indicate that after a limited number of epochs, the distribution $p(m)$ reaches a steady state, enabling the Gaussian mixture model to provide a stable and optimal fit for $p(m)$.

\begin{figure}[t]
	\centering
	\includegraphics[width=1.0\linewidth]{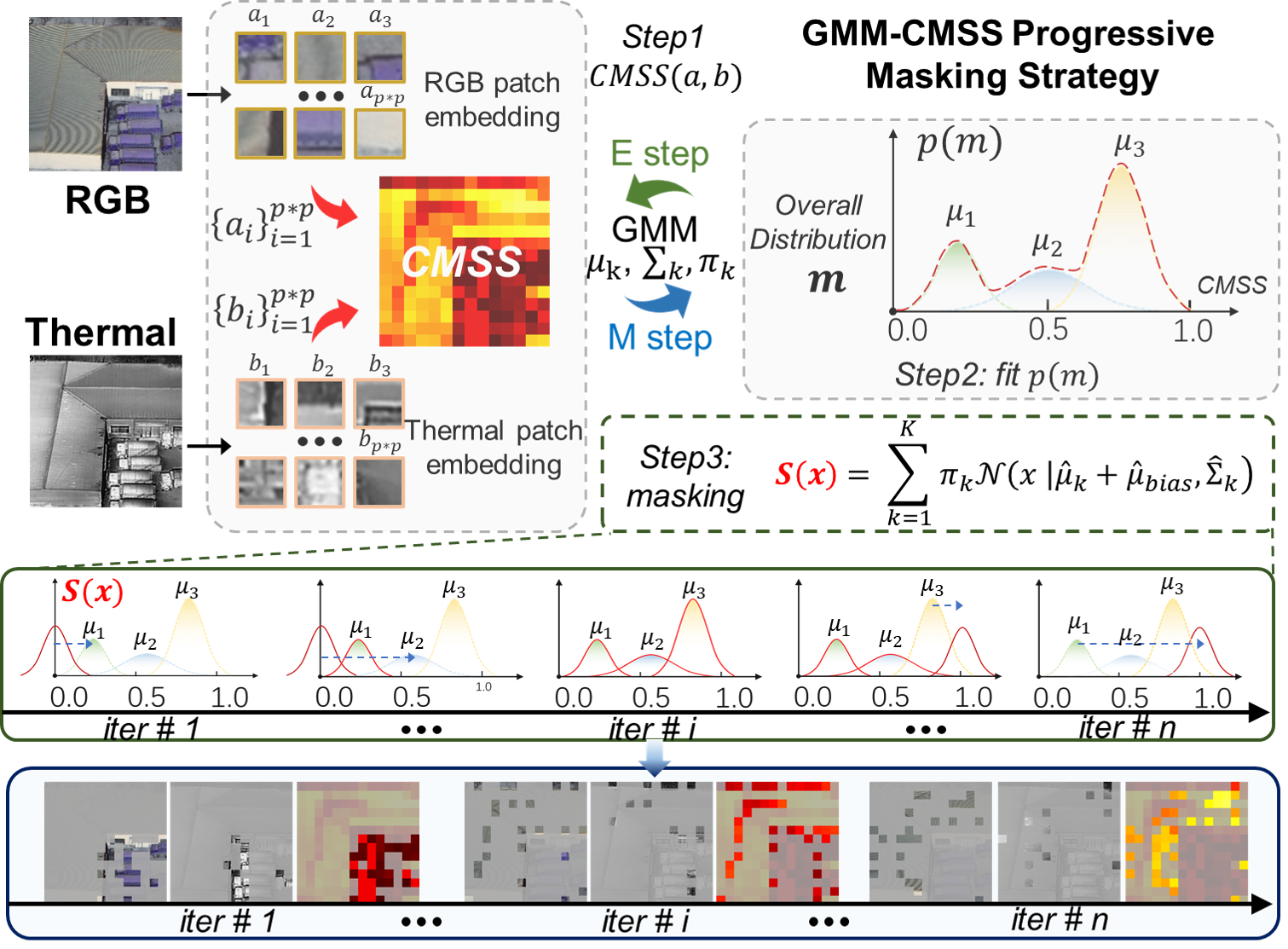}
	\vspace{-2.0em}
	\caption{As the sampling function $S(x)$ shifts from  $\hat{\mu} = 0$ to $\hat{\mu} = 1$ (green box), unmasked patches transition  from high- to low-information-density areas (blue box).}
	\label{fig:finetune}
 \vspace{-1.2em}
 \label{fig:rebutal_fig1}
\end{figure}

\subsection{GMM-CMSS Progressive Masking Strategy}
After approximating the $p(m)$ with the Gaussian mixture model parameters $\left\{ \mu_k, \Sigma_k, \pi_k \right\}$, we propose  the GMM-CMSS progressive masking strategy, in which the sampling function $S(x)$ is defined as follows:

\vspace{-1.2em}
\begin{equation}
S(x)=\sum_{k=1}^K \pi_k \mathcal{N}\left(x \mid \hat{\mu}_k+\hat{\mu}_{\text {bias }}, \hat{\Sigma}_k\right), K = 1,2,...
\end{equation}
\vspace{-0.6em}

\noindent here, $\hat{\mu}_k$, $\hat{\Sigma}_k$ represent the mean and variance of $k$-th Gaussian sampling model, respectively, while $\hat{\mu}_{\text {bias }}$ denotes the mean sampling bias for modality balance. Specifically, in each pre-training iteration, a batch of $B$ image pairs contains $B \times p \times p$ image embeddings, the sampling function $S(x)$ generates $B \times p  \times p$  sampling points $\textbf{s} =\left\{x_i\right\}_{i=1}^{B \times p \times p}$. For the CMSS distribution $\textbf{m}_{iter} =\left\{ m_i\right\}_{i=1}^{B \times p \times p}$ of the current iteration, we sample $B\times p\times p\times (r+r_{bias})$ masked patches from $\textbf{m}_{iter}$ that are nearest to the generated sampling points $\textbf{s}$, where $r$ is the masking ratio and $r_{bias}$ is a bias adaptively adjusted based on the modality loss difference. As illustrated in Fig.~\ref{fig:finetune}, we achieve the progressive masking strategy through controlling  the parameters $K$, $\hat{\mu}_k$ and $\hat{\Sigma}_k$. At the beginning of pre-training, we  initialize the sampling function $S(x)$ with $K = 1$, $\hat{\mu}_1 = 0$, and $\hat{\Sigma}_1 = 0.01$, ensuring unmasked patches are concentrated in high-information-density regions. As pre-training progresses, we gradually increase $\hat{\mu}_1$ from $0$ to  $\mu_1$, and the intermediate variance $\hat{\Sigma}_1$ is obtained through bilinear interpolation. Once $\hat{\mu}_1 = {\mu}_1$, we update the sampling function $S(x)$ with an additional Gaussian component, setting  $K = 2$,  $\hat{\mu}_2 = 0$, and $\hat{\Sigma}_2 = 0.01$. We implement the same operation for $\hat{\mu}_2$, $\hat{\Sigma}_2$ with $\hat{\mu}_1$, $\hat{\Sigma}_1$. At the middle of training, the parameter configuration is $K$=3 and $\{\hat{\mu}_k = {\mu}_k,\hat{\Sigma}_k = {\Sigma}_k\}_{k=1,2,3}$. Under this setting, the sampling function $S(x)$ closely approximates the probability density function $p(m)$ of the overall CMSS distribution, which can be considered as the random masking. At the end of pre-training, we gradually adjust the parameters $\{\hat{\mu}_k = \mu_k\}_{k=1,2,3}$ to the $\{\hat{\mu}_k = 1.0\}_{k=1,2,3}$, one by one. This adjustment shifts the unmasked patches toward regions with lower information density.









Our GMM-CMSS progressive masking strategy offers the following advantages: 1) Lightweight: The additional computational cost required during pre-training is negligible. 2) Object-centric: Regions with high information density will receive more attention in the early stages of pre-training. 3) Progressive sampling: Our proposed strategy moves from high- to low-information-density regions, facilitating  an easy-to-hard self-supervised learning process.



\subsection{M-SpecGene for Downstream Tasks}
\label{sec:fine-tune}

Fig.~\ref{fig:method}(b) illustrates the fine-tuning of M-SpecGene on downstream tasks. First, RGB and thermal images are patchified into feature embeddings $\mathcal{F}_{rgb}, \mathcal{F}_{t} \in \mathbb{R}^{B \times C \times HW}$. $F_{rgb}$ and $F_{t}$ are concatenated along the batch dimension to form $\mathcal{F}_{rgbt} \in \mathbb{R}^{2B \times C \times HW}$. Next, $F_{rgbt}$ is processed in parallel by the M-SpecGene ViT encoder, which owns the capability to represent both RGB and thermal modalities. To fuse multispectral features in a simple way, the output feature $\mathcal{F}^{out}_{rgbt} \in \mathbb{R}^{2B \times C \times HW}$ of M-SpecGene is reshaped to $\mathcal{F}^{out}_{rgbt} \in \mathbb{R}^{B \times 2C \times H \times W}$. Finally, $F^{out}_{rgbt}$ is fed into the downstream task heads for detection (ViTDet \cite{li2022exploring}), segmentation (UperNet \cite{xiao2018unified}) or matching (LoFTR \cite{sun2021loftr}). This workflow provides new insights into multispectral fusion with two key advantages: 1) The straightforward fusion strategy leverages the capability of foundation model to eliminate the design of complex handcrafted modules; 2) RGB-based single-modality methods can be seamlessly adapted to RGBT two-modality tasks without extra modification. 


%





\begin{table*}[ht]
    \setlength{\abovecaptionskip}{0.0em}
    \setlength{\belowcaptionskip}{0.0em}
   
    \hspace{-0.019\textwidth}
    \begin{subtable}[t]{0.6\textwidth}
        \centering
        \footnotesize{
            \setlength{\tabcolsep}{1.2pt}{
                \begin{tabular}{c|cccccc|ccc}
                    \hline
                    \noalign{\hrule height 0.5pt}
                    Methods           & Near  & Medium & Far   & None  & Partial & Heavy & Day   & Night & All   \\ \hline
                    ACF  \cite{7298706}             & 28.74 & 53.67  & 88.20 & 62.94 & 81.40   & 88.08 & 64.31 & 75.06 & 67.74 \\
                    Halfway Fusion \cite{2016arXiv161102644L}   & 8.13  & 30.34  & 75.70 & 43.13 & 65.21   & 74.36 & 47.58 & 52.35 & 49.18 \\
                    IATDNN+IASS \cite{GUAN2019148}& 0.04  & 28.55  & 83.42 & 45.43 & 46.25   & 64.57  & 49.02 & 49.37 & 48.96 \\
                    CLAN    \cite{ZHANG201920} & 3.71  & 19.04  & 55.82 & 30.31 & 41.57   & 62.48 & 36.02 & 32.38 & 35.53 \\
                    MSDS-R-CNN  \cite{2018arXiv180804818L}      & 1.29  & 16.19  & 63.73 & 29.86 & 38.71   & 63.37 & 32.06 & 38.83 & 34.15 \\
                    AR-CNN  \cite{2019arXiv190102645Z}& 0.00  & 16.08  & 69.00 & 31.40 & 38.63   & 55.73 & 34.36 & 36.12 & 34.95 \\
                    MBNet  \cite{zhou2020improving} & 0.00  & 16.07  & 55.99 & 27.74 & 35.43   & 59.14 & 32.37 & 30.95 & 31.87 \\
                    TSFADet  \cite{yuan2022translation}  & 0.00  & 15.99  & 50.71 & 25.63 & 37.29   & 65.67 & 31.76 & 27.44 & 30.74 \\
                    CMPD  \cite{li2022confidence}  & 0.00  & 12.99  & 51.22 & 24.04 & 33.88   & 59.37 & 28.30 & 30.56 & 28.98 \\
                    CAGTDet \cite{yuan2024improving} & 0.00  & 14.00  & 49.40 & 24.48 & 33.20   & 59.35 & 28.79 & 27.73 & 28.96 \\
                    C2Former  \cite{yuan2024c}        & 0.00  & 13.71  & 48.14 & 23.91 & 32.84   & 57.81 & 28.48 & 26.67 & 28.39 \\
                    RSDet  \cite{zhao2024removal}  & 0.00  & 12.13  & 39.80  & 20.49 & 33.25   & 57.60  & 25.83 & 26.48 & 26.02 \\  
                    UniRGB-IR (ViT-B) \cite{yuan2024unirgb}  & 0.00  & 13.44  & 38.21 & 20.26 & \textbf{31.67}   & \textbf{55.03} & 25.93 & 23.95 & 25.21 \\ \hline
                    M-SpecGene (ViT-S) & 0.03 &16.00  & 40.54 & 22.70  &  33.92 & 55.91 & 28.28 &  25.15 & 27.28 \\
                    M-SpecGene (ViT-B) & \textbf{0.00}  & \textbf{12.05}  & \textbf{34.57} & \textbf{18.20} & 33.32   & 55.85 & \textbf{25.66} & \textbf{19.42} & \textbf{23.74} \\ \hline
\noalign{\hrule height 0.5pt}
                \end{tabular}
            } 
            \vspace{-0.25 em}
           \caption{Comparison results on nine test subsets of the  KAIST dataset in terms of $MR^{\text{-}2}$.}
        }
    \end{subtable}
    \hspace{-0.048\textwidth}
    \begin{subtable}[t]{0.5\textwidth}
        \centering
        \footnotesize{
            \setlength{\tabcolsep}{0.2pt}{
                \begin{tabular}{c|ccc|ccc}
                    \hline
                    \noalign{\hrule height 0.5pt}
                    \multirow{2}{*}{Methods} & \multicolumn{3}{c|}{FLIR} & \multicolumn{3}{c}{LLVIP}                                                         \\ \cline{2-7}
                                              & mAP                       & $\text{mAP}_{50}$                     & $\text{mAP}_{75}$       & mAP         & $\text{mAP}_{50}$       & $\text{mAP}_{75}$       \\ \hline
                    Halfway Fusion   \cite{2016arXiv161102644L}            & 35.8                      & 71.5                      & - & 55.1        & 91.4        & - \\
                    GAFF  \cite{zhang2021guided}                   & 37.4                      & 74.7                      & 31.3        & 55.8        & 94.0        & 60.2        \\
                    PronEn    \cite{chen2022multimodal}               & 37.9                      & 75.5                      & 31.8        & 51.5        & 93.4        & 50.2        \\
                    CSAA    \cite{cao2023multimodal}                 & 41.3                      & 79.2                      & 37.4        & 59.2        & 94.3        & 66.6        \\
                    CALNet     \cite{10.1145/3581783.3612651}              & -               & -               & - & 63.9        & - & - \\
                    TIRDet    \cite{10.1145/3581783.3613849}               & 44.3                      & 81.4                      & \textbf{41.1}       & 64.2        & 96.3        & 73.1        \\
                    MMI-Det   \cite{10570450}               & 40.5                      & 79.8                      & 35.8        & 64.4        & \textbf{98.9}        & 73.5        \\
                    GFL-Res50    \cite{li2020generalized}            & 44.0                      & 78.1                      &     -       &    -       &      -     &   -        \\
                    ICAFusion   \cite{shen2024icafusion}             & 41.4                      & 79.2                      & 36.9        & - & - & - \\
                    CrossFormer    \cite{lee2024crossformer}          & 42.1                      & 79.3                      & 38.5        & 65.1        & 97.4        & 75.4        \\
                    RSNet     \cite{zhao2024removal}         & 41.4                      & 81.1                      & - & 59.2        & 94.3        & - \\
                    UniRGB-IR (ViT-B)  \cite{yuan2024unirgb}    & 44.1                    & 81.4                      & 40.2        & 63.2        & 96.1        & 72.2        \\   \hline
                    M-SpecGene (ViT-S)        & 43.7                      & 82.4                      & 39.4        & 63.4        & 96.3        & 74.1        \\
                    M-SpecGene (ViT-B)        & \textbf{44.7}                      & \textbf{84.8}                      & 40.1        & \textbf{65.3}        & 97.4        & \textbf{75.4}        \\ \hline
                    \noalign{\hrule height 0.5pt}
                \end{tabular}
            }
            \vspace{-0.25 em}
             \caption{Evaluation on the FLIR and LLVIP datasets in terms of mAP.}
        }
    \end{subtable}
    \vspace{-0.1em}
\caption{Evalution of the proposed M-SpecGene on the KAIST, FLIR and LLVIP datasets for the multispectral object detection task. }
\vspace{-0.75em}
\label{tab1}
\end{table*}

\begin{table*}[t]
    \setlength{\abovecaptionskip}{0.0em}
    \setlength{\belowcaptionskip}{0.5em}
   
    \hspace{-0.008\textwidth}
    \begin{subtable}[t]{0.65\textwidth}
        \centering
        \footnotesize{
            \setlength{\tabcolsep}{1.55pt}{
                \begin{tabular}{c|ccccccccccccc|c}
                \hline \noalign{\hrule height 0.5pt}
                        & Bkg & Bike & Bicyclist & Car & Tricycle & Box & Pole & Curve & Person & mIoU (\%) \\ \hline
                PSTNet \cite{shivakumar2020pst900} & 95.03 & 62.25 & 58.48 & 85.41 & 44.18 & 83.00 & 71.65 & 62.15 & 72.21 & 67.98                         \\
                MFNet \cite{ha2017mfnet} &  96.31 & 65.87 & 64.07 & 89.70 & 62.10 & 83.93 & 77.14 & 66.18 & 80.29 & 74.08 \\
                RTFNet \cite{sun2019rtfnet}  & 96.40 & 67.96 & 67.41 & 90.39 & 65.96 & 85.91 & 78.02 & 67.22 & 78.90 & 75.48       \\
                EGFNet \cite{zhou2022edge} & 96.57 & 71.26 & 70.86 & 90.52 & 71.51 & 85.41 & 76.49 & 66.92 & 83.74 & 77.44     \\
                ECM \cite{ji2023semanticrt} & 96.55 & 75.04 & 75.50 & 90.26 & 74.01 & 85.61 & 77.23 & 68.28 & 85.02 & 79.26 \\
                UniRGB-IR (ViT-B)  \cite{yuan2024unirgb} & 96.33 & 68.72 & 64.79 & 90.33 & 69.43 & 85.57 & 76.44 & 65.56 & 79.79 & 75.21 \\  \hline
                M-SpecGene (ViT-S) & 96.74 & 73.82 & 71.17 & 91.01 & 73.08 & 85.87 & 77.95 & 68.51 & 84.64 & 78.42 \\
                M-SpecGene (ViT-B) & \textbf{96.81} & \textbf{75.99} & \textbf{75.51} & \textbf{91.11} & \textbf{76.79} & \textbf{86.05} & \textbf{78.41} & \textbf{68.64} & \textbf{85.66} & \textbf{79.84} \\ \hline \noalign{\hrule height 0.5pt}
                \end{tabular}
            }
        }
         \vspace{-0.25em}
        \caption{Quantitative segmentation results on each class of the SemanticRT test set.}
    \end{subtable}
    \hspace{0.02\textwidth}
    \begin{subtable}[t]{0.35\textwidth}
        \centering
        
        \footnotesize{
            \setlength{\tabcolsep}{2pt}{
                \begin{tabular}{c|c|c}
                        \hline  \noalign{\hrule height 0.5pt}
        Methods                & Backbone       & mIoU (\%) \\ \hline
        OCRNet    \cite{yuan2020object}   & ResNet-50      & 52.38    \\
        LMANet   \cite{paul2021local}   & ResNet-50      & 52.73    \\
        DeepLabv3+   \cite{chen2018encoder}      & ResNet-50      & 51.59    \\
        $\text{MVNet}_{DeepLabv3+}$ \cite{ji2023multispectral}  & ResNet-50      & 54.52    \\ 
        DPLNet   \cite{dong2023efficient}             & MiT-B5         & 57.90    \\  
        UniRGB-IR (ViT-B) \cite{dong2023efficient}             & ViT-B         & 56.46    \\  \hline
        M-SpecGene (ViT-S)     & ViT-S          & 60.49    \\
        M-SpecGene (ViT-B)     & ViT-B          & \textbf{63.02}    \\ 
        \hline \noalign{\hrule height 0.5pt}
                \end{tabular}
            }
        }
        \vspace{-0.20em}
        \caption{Quantitative evaluation on the MVSeg dataset.}
    \end{subtable}
    \vspace{-0.65em}
        \caption{Comparison of the M-SpecGene on the SemanticRT and MVSeg  datasets for the multispectral semantic segmentation task.}
        \vspace{-1.6em}
    \label{tab2}
\end{table*}

\section{Experiments}

\begin{table*}[t]
\footnotesize
\centering
\hspace{-0.04\textwidth}
\begin{minipage}{0.39\textwidth}
    \centering
    \setlength{\tabcolsep}{1.2pt}
    \begin{tabular}{c|cccc|c}
        \hline \noalign{\hrule height 0.5pt}
        \rule{0pt}{2ex}   & Person & Truck  & Vege. & Pole & mIoU (\%) \\ \hline
        SegMiF   \cite{liu2023multi}   & 65.5   & 42.4     & 85.1       & 35.7 & 58.5 \\
        MDRNet+  \cite{zhao2023mitigating}  & 67.0   & 27.0    & 82.7       & 45.3 & 55.5 \\
        SGFNet  \cite{wang2023sgfnet}   & 67.2   & 34.6     & 82.7       & 42.8 & 56.0 \\
        MRFS  \cite{zhang2024mrfs}    & \textbf{71.3}   & 34.4    & \textbf{87.0}       & \textbf{53.6} & \textbf{61.2} \\ 
         UniRGB-IR (ViT-B) \cite{yuan2024unirgb}  & 66.5   & 36.3    & 85.6      & 42.1 & 59.8 \\  \hline
        M-SpecGene (ViT-S)  & 68.8   & 22.6   & 86.2       & 50.0 & 56.5 \\
        M-SpecGene (ViT-B)  & 65.6   & \textbf{44.4}     & 86.9       & 52.8 & 60.1 \\ \hline \noalign{\hrule height 0.5pt}
        \end{tabular}
    \vspace{-1.2em}
    \caption{Evaluation on the FMB segmentation dataset.}
    \label{tab:3}
\end{minipage}%
\hspace{0.005\textwidth}
\begin{minipage}{0.345\textwidth}
    \centering
    \setlength{\tabcolsep}{0.8pt}
    \begin{tabular}{c|c|ccc}
            \hline \noalign{\hrule height 0.5pt}
             \rule{0pt}{2ex}  & Methods   & @$3^{\circ}$$\uparrow$ & @$5^{\circ}$$\uparrow$ & @$10^{\circ}$$\uparrow$ \\ \hline
            \multirow{4}{*}{\begin{tabular}[c]{@{}c@{}}Detector-\\based\end{tabular}}    & RIFT \cite{li2019rift}& 0.0 & 0.0 & 0.0\\
             & POS-GIFT \cite{hou2024pos}& 0.0 & 0.0 & 0.4\\
             & ReDFeat \cite{deng2022redfeat} & 0.0 & 0.0 & 0.0  \\
            & SP+LG  \cite{lindenberger2023lightglue}  & 1.1 & 8.4 & 16.2 \\ \hline
            \multirow{3}{*}{\begin{tabular}[c]{@{}c@{}}Detector-\\free\end{tabular}}  & SemLA \cite{xie2023semantics}  & 0.0 & 0.2 & 1.2 \\
            & LoFTR  \cite{sun2021loftr}  & 18.8 & 29.7 & 46.2 \\ \cline{2-5}
            & \raisebox{-0.2ex}{Ours (ViT-S)} & \raisebox{-0.2ex}{\textbf{20.5}} & \raisebox{-0.2ex}{\textbf{31.7}} & \raisebox{-0.2ex}{\textbf{48.2}}   \\ \hline \noalign{\hrule height 0.5pt}
        \end{tabular}
    \vspace{-1.2em}
    \caption{RGBT feature matching evaluation.}
    \label{tab:4}
\end{minipage}%
\hspace{-0.013\textwidth}
\begin{minipage}{0.24\textwidth}
    \setlength{\tabcolsep}{1.3pt}
    \centering
    \begin{tabular}{c|cccc}
    \hline \noalign{\hrule height 0.5pt}
       & $\text{E}_{\xi}^{\text{max}}$$\uparrow$ & $\text{F}_{\beta}^\text{max}$$\uparrow$  & $\text{S}_\alpha$ $\uparrow$ & MAE$\downarrow$   \\ \hline
    MGFL \cite{huang2021multi} & 0.822 & 0.727 & 0.745 & 0.084 \\
    MIDD  \cite{tu2021multi} & 0.928 & 0.859 & 0.867 & 0.049 \\
    CGFNet \cite{9493207} & 0.927 & 0.870 & 0.865 & 0.042 \\
    ADF  \cite{tu2022rgbt}  & 0.892 & 0.815 & 0.830 & 0.074 \\
    MGAI \cite{song2022multiple}  & 0.940 & \textbf{0.879} & 0.881 & 0.038 \\ \hline
    Ours (ViT-S)  & 0.847 & 0.722 & 0.781 & 0.081 \\
    Ours (ViT-B)  & \textbf{0.942} & 0.877 & \textbf{0.888} & \textbf{0.033} \\ \hline \noalign{\hrule height 0.5pt}
    \end{tabular}
    \vspace{-1.2em}
   
    \caption{Test on VI-RGBT1500.}
    \label{tab:5}
\end{minipage}
   
\end{table*}

\begin{table*}[]
\setlength{\tabcolsep}{5.335pt} 
    \vspace{-1.2em}
    \centering
    \footnotesize{
    \begin{tabular}{c|cccc|cccc|cccc}
    
        \hline \noalign{\hrule height 0.5pt}
        \multirow{2}{*}{Methods} & \multicolumn{4}{c|}{VT821} & \multicolumn{4}{c|}{VT1000} & \multicolumn{4}{c}{VT5000}                                                                                                                                                       \\ \cline{2-13}
                               & $S$$\uparrow$  & $adpE$$\uparrow$ & $adpF$$\uparrow$ & $MAE$$\downarrow$ & S$\uparrow$ &$adpE$$\uparrow$ & $adpF$$\uparrow$ & $MAE$$\downarrow$ & $S$$\uparrow$ & $adpE$$\uparrow$ & $adpF$$\uparrow$ & $MAE$$\downarrow$ \\ \hline
        S2MA \cite{liu2020learning}   & 0.811                      & 0.813                       & 0.709                      & 0.098           & 0.918       & 0.912          & 0.848          & 0.029           & 0.853       & 0.864          & 0.743          & 0.053           \\
        JLDCF \cite{fu2021siamese}    & 0.839                      & 0.830                       & 0.726                      & 0.076           & 0.912       & 0.899          & 0.829          & 0.030           & 0.861       & 0.860          & 0.739          & 0.050           \\
        MTMR \cite{wang2018rgb}   & 0.725                      & 0.815                       & 0.662                      & 0.109           & 0.706       & 0.836          & 0.715          & 0.119           & 0.680       & 0.795          & 0.595          & 0.114           \\
        FMSF \cite{zhang2019rgb}  & 0.760                      & 0.796                       & 0.640                      & 0.080           & 0.873       & 0.899          & 0.823          & 0.037           & 0.814       & 0.864          & 0.734          & 0.055           \\
        MIDD \cite{tu2021multi}   & 0.871                      & 0.895                       & 0.803                      & 0.033           & 0.915       & 0.933          & 0.880          & 0.027           & 0.868       & 0.896          & 0.799          & 0.043           \\
        ADF \cite{tu2022rgbt}   & 0.810                      & 0.842                       & 0.717 & 0.077 & 0.910 & 0.921 & 0.847      & 0.034  & 0.864 & 0.891& 0.778 & 0.048           \\
        LSNet \cite{zhou2023lsnet} & 0.877 & 0.911 & 0.827 & 0.033 & 0.924 & 0.936          & 0.887 & 0.022 & 0.876 & 0.916 & 0.827 & 0.036           \\ 
        UniRGB-IR (ViT-B)   \cite{yuan2024unirgb}   & 0.881 & 0.895 & 0.806 & 0.039           & \textbf{0.939} & 0.943 & 0.894 & 0.018 & \textbf{0.906} & \textbf{0.935} & 0.849  & \textbf{0.027}    \\  \hline
        M-SpecGene (ViT-S) & 0.783 & 0.826 & 0.703 & 0.079 & 0.867  & 0.889 & 0.827 & 0.043 & 0.853 & 0.892 & 0.803 & 0.044 \\
        M-SpecGene (ViT-B) & \textbf{0.891} & \textbf{0.919} & \textbf{0.862} & \textbf{0.028} & 0.935 & \textbf{0.952} & \textbf{0.925} & \textbf{0.015} & 0.892 & 0.928 & \textbf{0.872} & 0.028 \\ \hline \noalign{\hrule height 0.5pt}
    \end{tabular}
    }
    \vspace{-1.2em}
    \caption{Comparison of M-SpecGene on the VT821, VT1000 and VT5000  datasets for the multispectral salient object detection task.}
    \vspace{-1.2em}
      \label{tab:6}
\end{table*}

\subsection{Implementation Details}
To maximize the utility of available unimodal and aligned RGBT data, M-SpecGene is first pre-trained on ImageNet \cite{deng2009imagenet} and single-modality thermal datasets to initialize the encoder and two decoders. Subsequently, M-SpecGene is further pretrained on the RGBT550K dataset to promote consistent representation. The RGB and thermal images undergo same preprocessing, including cropping within a range of 0.2x to 1.0x and a 50\% probability of random flipping. By default, a 90\% masking ratio is applied to both RGB and thermal images initially, and the AdamW optimizer is used with a base learning rate of ${1.5 \times  10}^{-4}$ and a half-cycle cosine decay schedule on 8 GTX 4090 GPUs. Following previous studies \cite{hong2024spectralgpt,liu2025infmae,LiSARATRX25}, after self-supervised pre-training, M-SpecGene is full-parameter fine-tuned on downstream RGBT multispectral tasks.

\begin{table*}[t]
    \setlength{\abovecaptionskip}{0.0em}
    \setlength{\belowcaptionskip}{0.5em}
    \hspace{-0.042\textwidth}
    \begin{subtable}[b]{0.45\textwidth}
        \centering
        \footnotesize
        \setlength{\tabcolsep}{1pt}
        \begin{tabular}{c|ccc|ccc}
        \hline \noalign{\hrule height 0.5pt}
        Methods& mAP  & $\text{mAP}_{50}$ & $\text{mAP}_{75}$ & $@3^{\circ}$$\uparrow$ & $@5^{\circ}$$\uparrow$  & @$10^{\circ}$$\uparrow$ \\ \hline
        From Scratch    & 36.0 & 70.6  & 32.0  & 12.5 & 23.6 & 41.2 \\
        Sup. (IN1K)           & 40.6 & 79.3  & 34.0  & 12.5 & 23.3 & 40.3 \\
        MAE (IN1K)           & 43.0 & 82.8  & 37.8  & 8.4 & 18.7 & 37.0 \\ \hline
        M-SpecGene& 44.7 & 84.8  & 40.1  & 20.5 & 31.7 & 48.2 \\ \hline \noalign{\hrule height 0.5pt}
        \end{tabular}
        \caption{Comparisons on different pretrained models.}
    \end{subtable}
    \hspace{-0.033\textwidth}
    \begin{subtable}[b]{0.15\textwidth}
        \centering
        \footnotesize
        \setlength{\tabcolsep}{0.50mm}
        \begin{tabular}{c|c}
        \hline \noalign{\hrule height 0.5pt}
        Architecture & $\text{mAP}_{50}$ \\ \hline
        Vanilla MAE  & 83.1 \\
         Concat & 80.1 \\
        Auxiliary  & 83.5 \\
        Siamese & 83.8 \\
        \hline \noalign{\hrule height 0.5pt}
        \end{tabular}
        \caption{Architecture.}    
    \end{subtable}
    \hspace{0.006\textwidth}
    \begin{subtable}[b]{0.13\textwidth}
        \centering
        \footnotesize
        \setlength{\tabcolsep}{0.50mm}
        \begin{tabular}{c|c}
        \hline \noalign{\hrule height 0.5pt}
        Masking & $\text{mAP}_{50}$ \\ \hline
        Random  & 83.8 \\
        Low CMSS & 83.6 \\
        High CMSS & 83.4 \\
        GMM-CMSS & 84.8 \\
        \hline \noalign{\hrule height 0.5pt}
        \end{tabular}
        \caption{Masking way.}
    \end{subtable}
    \hspace{0.025\textwidth}
    \begin{subtable}[b]{0.13\textwidth}
        \centering
        \footnotesize
        \setlength{\tabcolsep}{5pt}
        \begin{tabular}{c|c}
        \hline \noalign{\hrule height 0.5pt}
        Blocks & $\text{mAP}_{50}$ \\ \hline
        2 & 84.1 \\ \hline
        4 & 84.8 \\ \hline
        8 & 84.5 \\
        \hline \noalign{\hrule height 0.5pt}
        \end{tabular}
        \caption{Decoder Depth.}
    \end{subtable}
    \hspace{0.01\textwidth}
    \begin{subtable}[b]{0.13\textwidth}
        \centering
        \footnotesize
        \setlength{\tabcolsep}{5pt}
        \begin{tabular}{c|c}
        \hline \noalign{\hrule height 0.5pt}
        Ratio & $\text{mAP}_{50}$ \\ \hline
        85\% & 84.4  \\ \hline
        90\% & 84.8 \\ \hline
        95\% & 84.1 \\
        \hline \noalign{\hrule height 0.5pt}
        \end{tabular}
        \caption{Masking ratio.}
    \end{subtable}
    \vspace{-1.8em}
    \caption{Ablation analysis of M-SpecGene in terms of pretrained model, architecture, masking strategy, decoder depth and masking ratio.}
    \vspace{-1.5em}
    \label{tab:7}
\end{table*}
\begin{figure*}[t]
	\centering
    
	\includegraphics[width=1.0\linewidth]{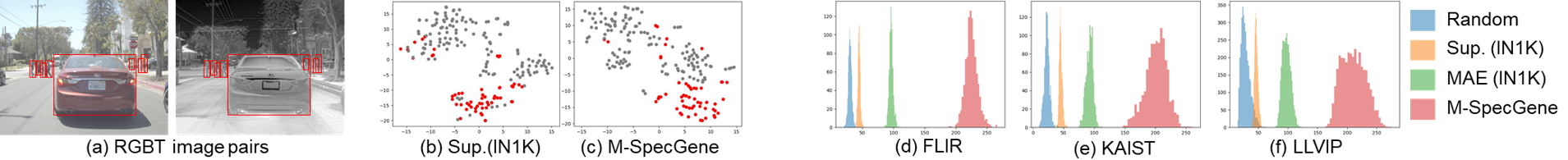}
	\vspace{-2.3em}
	\caption{(a) Samples for feature visualization. (b-c) The t-SNE visualization of concatenated RGBT features for object and background regions. (d-f) The statistical distribution of the Wasserstein distance between object and background features on three detection datasets.}
	\label{fig:exp}
 \vspace{-1.2em}
\end{figure*}

\subsection{RGBT Multispectral Object Detection}

\textbf{Experimental Settings:} We validate M-SpecGene on the multispectral object detection across three datasets: KAIST \cite{hwang2015multispectral}, LLVIP \cite{jia2021llvip}, and FLIR \cite{zhang2020multispectral}. We evaluate pedestrian detection on the KAIST dataset using the log-average Miss Rate  over false positives per image ($MR^{\text{-}2}$). For the LLVIP and FLIR datasets, we use mean Average Precision (mAP) for evaluation. To fully leverage the capabilities of the plain vision transformer, we use ViTDet \cite{li2022exploring} as the detector. Notably, RGB and thermal images undergo consistent data augmentation, and RGBT features are fused via simple  concatenation of the ViT encoder outputs.

\textbf{Results and Analyses:} As shown in Tab.~\ref{tab1}(a), our M-SpecGene achieves the best performance across the seven of the nine evaluation metrics on the KAIST dataset, outperforming the previous best method UniRGB-IR \cite{yuan2024unirgb} by 1.47\% on the “ALL” set. On the FLIR and LLVIP datasets, the ViT-S version of M-SpecGene achieves performance comparable to UniRGB-IR, while the ViT-B version demonstrates an enhanced ability to leverage foundational model strengths in Tab.~\ref{tab1}(b), achieving higher detection accuracy than previous methods. It should be noted that the ViT-B in UniRGB-IR is pretrained on COCO dataset first, while our M-SpecGene does not rely on the high-quality RGB detection dataset for extra improvement. With the learned self-supervised representation from large-scale data, our M-SpecGene can effectively fuse RGB and infrared modalities without complex handcrafted modules.

\subsection{RGBT Multispectral Semantic Segmentation}
\textbf{Experimental Settings:} Three recently released datasets which own high-quality samples are used for the validation on the multispectral semantic segmentation task. The SemanticRT \cite{ji2023semanticrt}, MVSeg \cite{ji2023multispectral} and FMB \cite{liu2023multi} datasets include 13, 26, and 15 categories, respectively. Mean Intersection over Union (mIoU) across all categories is used to evaluate semantic segmentation performance. Following MAE \cite{he2022masked}, we employ UperNet \cite{xiao2018unified} as the base segmentation framework. The model architecture remains unchanged and only a simple concatenation operation is added.

\textbf{Results and Analyses:} We compare M-SpecGene with competitive methods on the SemanticRT dataset in Tab.~\ref{tab2}(a) and the MVSeg dataset in Tab.~\ref{tab2}(b). Quantitative results confirm the effectiveness of M-SpecGene on both datasets. MVNet \cite{ji2023multispectral} serves as simple baseline that uses multispectral video clips to leverage extra temporal information, while M-SpecGene achieves higher mIoU accuracy by only utilizing the frame-level information. Tab.~\ref{tab:3} shows that on the FMB dataset, M-SpecGene is superior to other competitive methods but falls short of MSRS \cite{zhang2024mrfs} on certain metrics. Given that FMB is a small-scale dataset with only 280 validation samples, MSRS and UniRGB-IR, which incorporate complex fusion modules based on Segformer \cite{xie2021segformer}, tend to fit the FMB more easily than M-SpecGene, which only employs a simple concatenation operation for feature fusion.  M-SpecGene tends to achieve superior performance, particularly in scenarios involving extensive category diversity, large-scale datasets, and high task complexity.


\subsection{RGBT Cross-modality Feature Matching}
\textbf{Experimental Settings:} Considering the high alignment quality, LLVIP \cite{jia2021llvip} dataset is used to evaluate cross-modality feature matching. The Area Under the Curve (AUC) metric is used for evaluation. We adopt the widely recognized LoFTR \cite{sun2021loftr} as the basic framework, with the backbone replaced by ViT-S. To enhance locality, we incorporate a convolutional stem \cite{xiao2021early}.

\textbf{Results and Analyses:} Tab.~\ref{tab:4} shows that traditional handcrafted feature descriptors  struggle to handle complex scenes in the LLVIP dataset. Moreover, detector-based methods yield unsatisfactory results due to difficulties in extracting repeatable keypoints across two modalities. Our M-SpecGene significantly outperforms other methods at various thresholds, as the learned modality-invariant representation facilitates the RGBT feature matching with reduced modality characteristic differences in latent space.

\subsection{RGBT Multispectral Salient Object Detection}
\textbf{Experimental Settings:}  The VT821 \cite{wang2018rgb}, VT1000 \cite{tu2019rgb}, VT5000 \cite{tu2022rgbt} and VI-RGBT1500 \cite{song2022multiple} are used for evaluation on the multispectral salient object detection. F-measure ($adpF$, $\text{F}_{\beta}^\text{max}$), E-Measure ($adpE$, $\text{E}_{\xi}^{\text{max}}$), S-Measure ($S$) and Mean Absolute Error ($MAE$) are adopted as metrics. We employ the UperNet \cite{xiao2018unified} as the basic framework and follow the common setting that 2,500 image pairs in the VT5000 dataset are treated as the training dataset, while the remaining and other datasets are used as the test sets.

\textbf{Results and Analyses:} Experiments in  Tab.~\ref{tab:5} and Tab.~\ref{tab:6} show that M-SpecGene achieves better results than previous methods  across eleven subset metrics, with particularly notable improvements on the VT821, VT1000, and VI-RGBT1500 datasets, rather than the VT5000 dataset. This highlights its superior generalization capability.




\subsection{Ablation Study}
\textbf{Comparisons on Pretrained Models:} In Tab.~\ref{tab:7}(a), we compare the performance of different pretrained models in multispectral object detection using KAIST dataset and cross-modality feature matching on LLVIP dataset. We observe that ViT trained from scratch performs poorly in terms of mAP on FLIR. While vanilla MAE-pretrained ViT improves $\text{mAP}_{50}$ from 40.6\% to 43.0\% compared to the Supervised (Sup.) pretrained ViT. M-SpecGene exhibits superior performance by further improving the $\text{mAP}_{50}$ to 44.8\%. On the LLVIP dataset, M-SpecGene significantly boosts AUC@$10^{\circ}$ from 41.2 to 48.2, whereas both supervised and vanilla MAE pretrained ViT models lead to a decline in matching accuracy. We attribute this discrepancy to the inherent difference between detection and matching tasks. The detection task aims to leverage both modalities to generate complementary features, whereas the matching task focuses on identifying the common features shared by both modalities. Therefore, pre-training on the single-modality ImageNet dataset may disrupt symmetrical representations required for cross-modality feature matching. Overall, effective pre-training for modality-invariant representation is crucial for a generalized multispectral foundation model.



\textbf{RGBT Representation Architecture:} To investigate effective self-supervised representation architectures for both RGB and thermal modalities, we design four approaches: 1) Vanilla MAE \cite{he2022masked}: RGB and thermal images are mixed in the input level, and a vanilla MAE is employed. 2) Channel concatenation: RGB and thermal images are concatenated along the channel dimension. 3) Auxiliary branch: Complementary masked RGB and thermal patches are processed with a shared-weight encoder, then thermal features serve as auxiliary information in the cross-attention layer to aid the RGB decoder in reconstructing the masked region. 4) Siamese-based: RGB and thermal modalities are encouraged to learn consistent representations with a shared-weight encoder, with independent decoders applied to each modality. Tab.~\ref{tab:7}(b) shows the Siamese-based architecture achieves the best results, which reserves the symmetry and fully utilizes cross-modality complementarity.

\textbf{Masking Strategy:} We compare four different masking strategies in Tab.~\ref{tab:7}(c): 1) Random masking. 2) Gaussian masking in the low-CMSS region. 3) Gaussian masking in the high-CMSS region. 4) GMM-CMSS progressive masking. Experimental results indicate that focusing on a single information density region leads to inferior performance. In contrast, GMM-CMSS progressive masking enables a flexible, easy-to-hard, and object-centered learning process, thereby producing more robust representations.

\textbf{Decoder Depth:} Tab.~\ref{tab:7}(d) shows a decoder depth of four achieves the best results, indicating that the default decoder depth of MAE \cite{he2022masked} can be reduced under the Siamese-based architecture with two independent decoders.

\textbf{Masking Ratio:}  Tab.~\ref{tab:7}(e) illustrates that a lower masking ratio, which reduces the reconstruction difficulty particularly  for the thermal modality, leads to a decrease in mAP slightly. A higher masking ratio will negatively affect the effectiveness of the GMM-CMSS strategy. Therefore, we set the default masking ratio to 90\%.

\textbf{Feature Visualization and Statistical Analysis:} We first concatenate the RGB and thermal features extracted by pretrained models and perform a visual analysis of the concatenated object and background features. As shown in Fig.~\ref{fig:exp}(b-c), the object features extracted by M-SpecGene exhibit greater discriminability compared to those from the Sup. (IN1K) pretrained model. Subsequently, we conduct a statistical analysis of the differences  between object and background features across three  detection datasets. Specifically, we compute the Wasserstein distance between object and background features for each sample and present the statistical Wasserstein distance distribution of different pretrained models. Fig.~\ref{fig:exp}(d-f) show that the model trained from scratch exhibits smaller overall Wasserstein distances, whereas the distributions of MAE (IN1K) and Sup. (IN1K) show larger  Wasserstein distances. Notably, our M-SpecGene achieves the largest Wasserstein distance distribution, indicating more significant feature differences between objects and backgrounds. This suggests that the GMM-CMSS progressive masking strategy facilitates the learning of more object-centric representations, thereby promoting the generation of more discriminative features.

\section{Conclusion}


We make the first attempt to build a multispectral foundation model,  aiming to transform previous case-by-case studies  into a unified paradigm. To mitigate the impact of information imbalance  inherent in RGBT datasets, we introduce the CMSS metric to measure cross-modality information density and develop a GMM-CMSS progressive masking strategy to enable a flexible, easy-to-hard, and object-centric pre-training progress. The proposed M-SpecGene  effectively represents both RGB and thermal modalities in the latent space,  eliminating the need for handcrafted modules and offering new insights into multispectral fusion.  Extensive experiments on eleven datasets across four tasks validate the generalizability of M-SpecGene, which can fully expolit the carefully constructed, high-quality RGBT550K dataset for self-supervised pre-training and seamlessly adapt RGB single-modality methods to RGBT two-modality tasks without extra modification. We hope this work will advance the application of multispectral vision from the perspective of generalized foundation model. 


\clearpage
 \section*{Acknowledgments} 
 This research was supported by  National Science Fund for Distinguished Young Scholars (62025108). 
 In addition, this research was supported in part by the Agency for Science, Technology and Research (A\textsuperscript{\textasteriskcentered}STAR) under its IAF-ICP Programme I2501E0041 and the Schaeffler-NTU Corporate Lab (SHARE\texttt{@}NTU), and in part by the National Research Foundation Singapore Competitive Research Program (award number CRP29-2022-0003). The work was done at Rapid-Rich Object Search (ROSE) Lab, School of Electrical \& Electronic Engineering, Nanyang Technological University.
{
    \small
    \bibliographystyle{ieeenat_fullname}
    \bibliography{main}
}


\end{document}